\documentclass[10pt,twocolumn,letterpaper]{article}

\usepackage[pagenumbers]{wacv} 

\usepackage{bbm}
\usepackage{array,multirow}
\usepackage{arydshln}  
\usepackage{float}

\usepackage[utf8]{inputenc}  
\usepackage[T1]{fontenc}     

\definecolor{wacvblue}{rgb}{0.21,0.49,0.74}
\usepackage[pagebackref,breaklinks,colorlinks,allcolors=wacvblue]{hyperref}

\def\wacvPaperID{568} 
\def\confName{WACV}
\def\confYear{2026}

\title{SIDE: Sparse Information Disentanglement for Explainable Artificial Intelligence}

\author{Viktar Dubovik\textsuperscript{1}
\hspace{1.0em}    Łukasz Struski\textsuperscript{1}
\hspace{1.0em}    Jacek Tabor\textsuperscript{1}
\hspace{1.0em}    Dawid Rymarczyk\textsuperscript{1,2} \\ 
$^1$ Jagiellonian University, Faculty of Mathematics and Computer Science\\
$^2$ Ardigen SA\\
}

\begin{document}
\maketitle
\begin{abstract}
  
  
  Understanding the decisions made by deep neural networks is essential in high-stakes domains such as medical imaging and autonomous driving. 
Yet, these models often lack transparency, particularly in computer vision. 
Prototypical-parts-based neural networks have emerged as a promising solution by offering concept-level explanations. 
However, most are limited to fine-grained classification tasks, with few exceptions such as InfoDisent. 
InfoDisent extends prototypical models to large-scale datasets like ImageNet, but produces complex explanations.
  
  We introduce Sparse Information Disentanglement for Explainability (SIDE), a novel method that improves the interpretability 
of prototypical parts through a dedicated training and pruning scheme that enforces sparsity. 
Combined with sigmoid activations in place of softmax, this approach 
allows SIDE to associate each class with only a small set of relevant prototypes. Extensive experiments show that SIDE matches the accuracy of existing methods while reducing explanation size by over 90\%, 
substantially enhancing the understandability of prototype-based explanations.
\end{abstract}

\section{Introduction}

\begin{figure}[t!] 
\centering
\includegraphics[width=1\columnwidth]{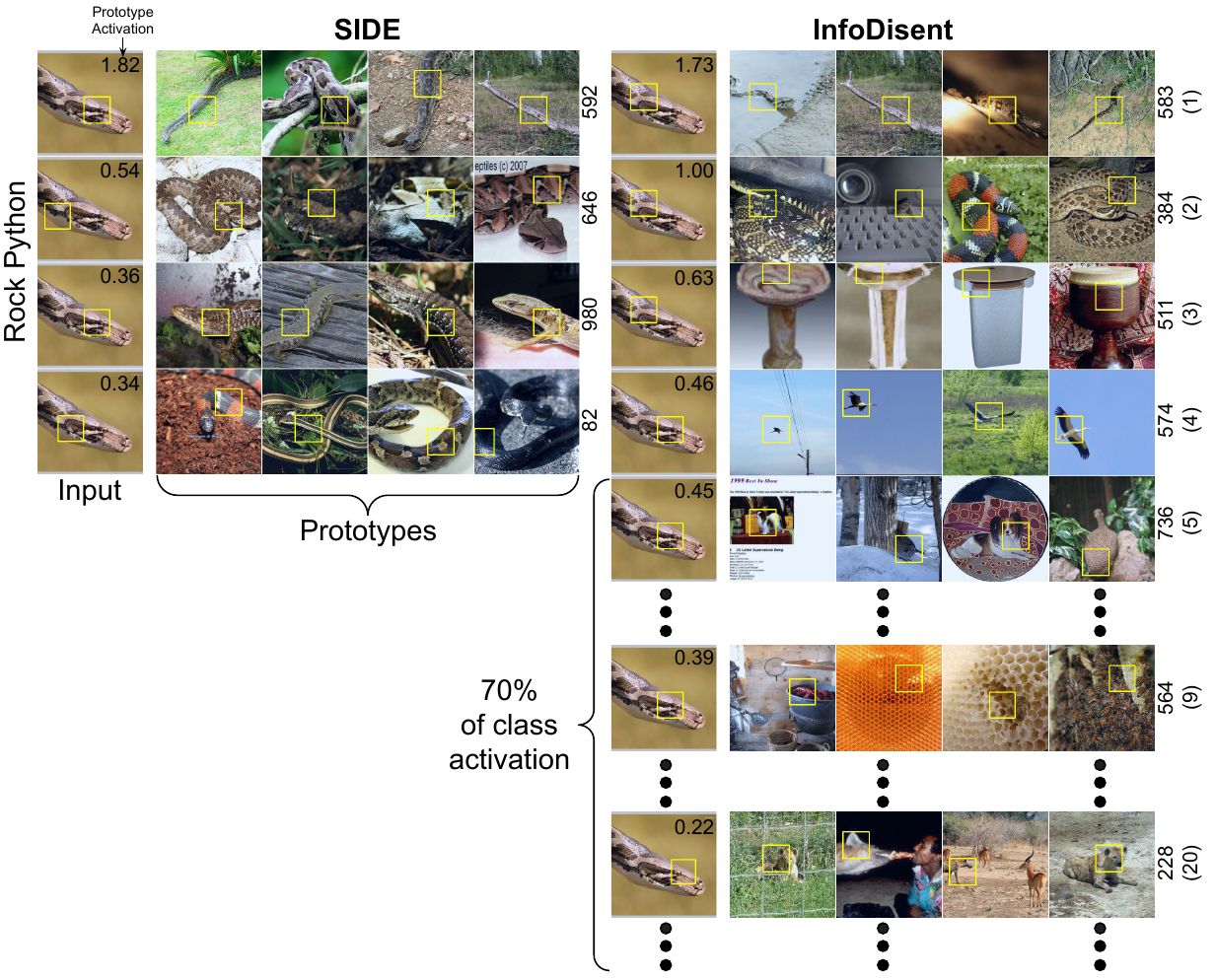}
\caption{Comparison of SIDE and InfoDisent prototypical explanations for a SwinV2-S model on ImageNet. Each row shows one prototype (channel index at right). In the leftmost column, the test image is overlaid with that prototype’s activation score (top-right) and the yellow box marking its activated region; the remaining columns display the top training patches that most strongly activate that prototype. SIDE produces a compact explanation—only 4 prototypes for this image (5 per class). InfoDisent activates 21 prototypes above the 0.2 threshold and 747 below (768 total), with its top 4 prototypes accounting for just 30\% of the class prediction score (total activation = 12.89).}
\label{fig:side}
\vspace{-2em}
\end{figure}

Deep Neural Networks (DNNs) excel in computer vision tasks, often surpassing human accuracy. However, their black-box nature make the adoption difficult in safety-critical domains such as medical diagnosis and autonomous driving \citep{buhrmester2019analysisexplainersblackbox, Lapuschkin_2019, rudin2021interpretablemachinelearningfundamental}. Explainability is thus crucial for technical validation, regulatory compliance, and building trust with domain experts \citep{Bibal2020, molnar2025}. To address this, intrinsically interpretable models, particularly concept-based approaches emerged such as ProtoPNet \cite{chen2019lookslikethatdeep}, offering higher-level explanations by identifying prototypical parts.

While improving interpretability, existing concept-based methods are largely limited to Convolutional Neural Networks (CNNs) and small, fine-grained datasets. Adapting them to Vision Transformers (ViTs) \citep{dosovitskiy2021imageworth16x16words} and large-scale datasets like ImageNet \cite{russakovsky2015imagenetlargescalevisual} is essential for broader applicability. Early attempts, such as ProtoPFormer \cite{xue2022protopformerconcentratingprototypicalparts}, required architecture-specific adaptations and training transformers from scratch, often underperforming baselines. InfoDisent~\cite{struski2024infodisent} aimed to bridge this gap by combining concept-based explanations with the pre-trained models, enabling its application to frozen CNN and ViT backbones on large-scale datasets. However, InfoDisent faces challenges: the number of learned concepts is limited by feature dimensionality, often insufficient for large datasets; predictions frequently activate hundreds of prototypes, harming interpretability; and its reliance on a standard Softmax classifier degrades interpretability~\cite{gairola2025probe}.

To overcome these limitations, we introduce \textbf{SIDE} (Sparse Information Disentanglement for Explainability) — a novel architecture extending the InfoDisent to produce more compact, interpretable explanations, and improved performance. These advancements are achieved through a new training procedure incorporating multi-label classification loss and weight pruning, alongside architectural modifications: increasing the prototype capacity, replacing the final interpretable linear layer with a sparse variant, and switching from softmax to sigmoid activations.

We validate SIDE through extensive experiments on standard XAI benchmarks (CUB \cite{wah_2023_cvm3y-5hh21}, Stanford Dogs \cite{KhoslaYaoJayadevaprakashFeiFei_FGVC2011}, Stanford Cars \cite{6755945}) and ImageNet \cite{russakovsky2015imagenetlargescalevisual}. Our results demonstrate that SIDE consistently outperforms other concept-based models with the same number of prototypes, achieving comparable or superior performance to InfoDisent. Notably, SIDE excels on ImageNet with the SwinV2 backbone, surpassing InfoDisent in accuracy while activating under 9 prototypes per prediction on average. These findings demonstrate SIDE's ability to provide compact and interpretable explanations across CNN and ViT architectures, scaling effectively to large-scale datasets like ImageNet.

Our contributions are summarized as follows:
\begin{itemize}
    \item We introduce SIDE, a novel concept-based, self-explainable neural network that provides compact and interpretable explanations due to the usage of sigmoid activations instead of softmax one, across both CNN and Vision Transformer architectures.
    \item We propose a joint training and pruning strategy that promotes sparsity in the learned concepts, resulting in concise and human-interpretable explanations.
    \item We extensively evaluate SIDE on diverse benchmarks, demonstrating its effectiveness on both fine-grained classification tasks and large-scale datasets such as ImageNet.
\end{itemize}

\section{Related Works}

The pursuit of interpretable AI models has advanced self-explainable (ante-hoc) methods, especially in computer vision~\citep{rudin2019stop, NEURIPS2018_3e9f0fc9, bohle2022b, brendel2018approximating}. A key direction is ante-hoc explainability focused on enhancing prototypical parts, a concept introduced by ProtoPNet~\citep{chen2019lookslikethatdeep} that uses learned visual prototypes to represent neural network activation patterns.

Subsequent research has expanded ProtoPNet in various ways. To achieve more distinct representations, TesNet~\citep{wang2021interpretable} and Deformable ProtoPNet~\citep{donnelly2022deformable} integrated orthogonality in prototype construction. For efficiency, methods like ProtoPShare~\citep{rymarczyk2021protopshare}, ProtoTree~\citep{nauta2021neural}, ProtKNN~\citep{ukai2022looks}, and ProtoPool~\citep{rymarczyk2022interpretable} reduced the number of required prototypes. Prototype-based methods have also been extended to hierarchical classification~\citep{hase2019interpretable} and knowledge distillation~\citep{keswani2022proto2proto}.

Recent efforts have also focused on improving the interpretability of prototypical explanations. This includes enhancing spatial alignment~\citep{sacha}, examining low-level feature attribution~\citep{nauta2021looks}, disentangling visual attributes by separately processing color information~\citep{pach2024lucidppn}, and enabling prototype disentanglement through interactive user feedback~\citep{michalski2025personalized}. The versatility of prototype-based solutions has led to their widespread adoption in diverse applications, including medical imaging~\citep{afnan2021interpretable, barnett2021case, kim2021xprotonet, rymarczyk2022protomil}, time-series analysis~\citep{gee2019explaining}, graph classification~\citep{rymarczyk2022progrest, zhang2021protgnn}, semantic segmentation~\citep{sacha2023protoseg}, and challenging learning paradigms like class incremental learning~\citep{rymarczyk2023icicle}.

More recently, prototypical parts have been generalized to modern deep learning architectures, including transformer models~\cite{ma2024interpretable}, and applied to large-scale datasets such as ImageNet with approaches like InfoDisent~\citep{struski2024infodisent} and EPIC~\citep{borycki2025epic}. Despite these advancements, a persistent challenge remains: models designed for large-scale datasets often struggle to provide equally plausible and intuitive explanations for finer-grained details. This work addresses this critical gap by introducing SIDE, a novel approach for generating compelling and understandable explanations for large-scale image classification.

\begin{figure}[t!] 
\centering
\includegraphics[width=0.95\columnwidth]{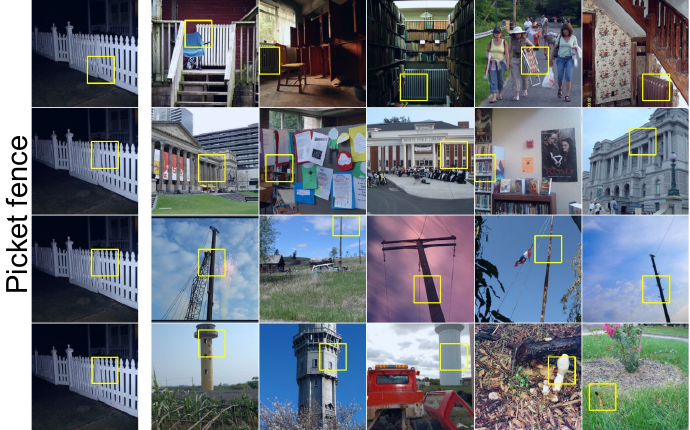}
\caption{SIDE explanation for the ImageNet class Picket Fence using SwinV2-S.
Prototypes 1–2 capture densely packed slats and regularly spaced vertical structures, reflecting the repetitive upright pattern of picket fences.
Prototypes 3–4 focus on isolated vertical elements, from narrow posts (P3) to thicker pillars and mushroom stems (P4).
Only 4 of the 9 learned prototypes are shown for clarity.}
\vspace{-2.1em}
\label{fig:main_imagenet}
\end{figure}

\section{Preliminaries}
The Preliminaries section introduces the two building blocks on which \textbf{SIDE} is based:  

(i) the InfoDisent \cite{struski2024infodisent} method for post-hoc information disentanglement (§\ref{sec:info-disent}), and  

(ii) the Asymmetric Loss (ASL) \cite{benbaruch2021asymmetriclossmultilabelclassification} for training in imbalanced multi-label settings (§\ref{subsec:asl}).

\subsection{InfoDisent}
\label{sec:info-disent}

InfoDisent is a concept-based interpretability method that enables prototypical part-based explanations on top of any pretrained image classifier. The method assumes a frozen feature extractor (CNN or Vision Transformer) and appends an interpretable classification head that disentangles information from the final feature maps of the model into semantic units, referred to as prototypes.

\textbf{Orthogonal disentanglement.}  
Let \(F\!\in\!\mathbb{R}^{d\times H\times W}\) be the final feature map tensor of the backbone (with \(d\) channels).  
A trainable orthogonal map \(U\!\in\!\mathbb{R}^{d\times d}\) is applied channel-wise:
\[
\tilde F = U F \in \mathbb{R}^{d\times H\times W}.
\]
This isometry enables disentanglement of meaningful independent concepts from \( F \).

\textbf{Sparse pooling.}  
For each channel \(K = \tilde F_c \in \mathbb{R}^{H\times W}\) a sparse information bottleneck is then computed to enforce the disentanglement of the channels:
\[
\operatorname{mxpool}(K) \;=\; \max\bigl(\operatorname{ReLU}(K)\bigr)
                               \;-\; \max\bigl(\operatorname{ReLU}(-K)\bigr).
\]
Stacking over all channels yields a pooled vector \(v\in\mathbb{R}^{d \times 1}\).

\textbf{Non-negative linear head.}  
Let \( W \in \mathbb{R}^{C \times d} \) denote the weight matrix of the final linear layer (Scores Sheet in short), which maps prototype presence scores \( v \) to unnormalized class logits \( z \in \mathbb{R}^{C} \). The final class probabilities distribution is denoted by \( p \in[0,1]^C \)

\( W \) weights are constrained to be non-negative; element-wise absolute value enforces this at training time:

\begin{equation}
p \;=\; \operatorname{softmax}\!\bigl(|W| v\bigr)\,,
\label{eq:infodisent_ll}
\end{equation}

By treating each channel as a standalone concept and enforcing positivity in the classification layer, InfoDisent enables explanations in the form of additive contributions of prototypical parts.

\subsection{Asymmetric Loss}
\label{subsec:asl}

\textbf{From BCE to Focal loss.}
For a logit \(z \in \mathbb{R}\), predicted probability \( p = \sigma(z) \), and binary label \(y \in \{0,1\}\), the Binary Cross-Entropy (BCE) loss is given by:
\[
\mathcal{L}_{\text{BCE}} = -y \log p - (1-y)\log(1-p).
\]
Focal loss~\cite{lin2018focallossdenseobject} was introduced to address the class imbalance problem by down-weighting easy examples. It adds a focusing term \((1-p)^{\gamma}\) to the loss:
\[
\mathcal{L}_{\text{FL}} = -y (1-p)^{\gamma} \log p - (1 - y) p^{\gamma} \log(1 - p),
\]
which emphasizes hard misclassified samples. While effective in object detection, focal loss is less suited for multi-label classification. In this setting, treating positive and negative samples symmetrically causes the loss to be dominated by frequent negatives, suppressing gradients from rare but crucial positive labels. Thus, the model overfits to negatives while under-learning discriminative features for positives.

\textbf{Asymmetric Loss}
To overcome this, Asymmetric Loss (ASL)~\cite{benbaruch2021asymmetriclossmultilabelclassification} decouples the treatment of positive and negative targets. It introduces separate focusing parameters \(\gamma_{+}\) and \(\gamma_{-}\), allowing strong supervision for positives to be preserved, while attenuating the contribution of easy negatives via \(\gamma_{+} < \gamma_{-}\). 

In addition, ASL applies a probability margin \(m \in [0,1)\) that shifts the predicted score for negative labels, effectively discarding samples where the predicted probability is very low. This hard thresholding further suppresses trivial negatives and mitigates the influence of potentially mislabeled examples. The combined formulation of ASL is:
\begin{equation}
\mathcal{L}_{\text{ASL}} =
\begin{cases}
(1 - p)^{\gamma_{+}} \log p, & y = 1, \\[4pt]
(p_m)^{\gamma_{-}} \log(1 - p_m), & y = 0,
\end{cases}
\label{eq:asl}
\end{equation}
where \(p_m = \max(p - m, 0)\). This design enables fine-grained control of gradient flow during training. 

Typical hyperparameters include \(\gamma_{+} = 0\), \(\gamma_{-} \in [2,4]\), and \(m \in [0.05, 0.2]\).

\section{SIDE}

\begin{figure*}[t!] 
\centering
\includegraphics[width=0.85\textwidth]{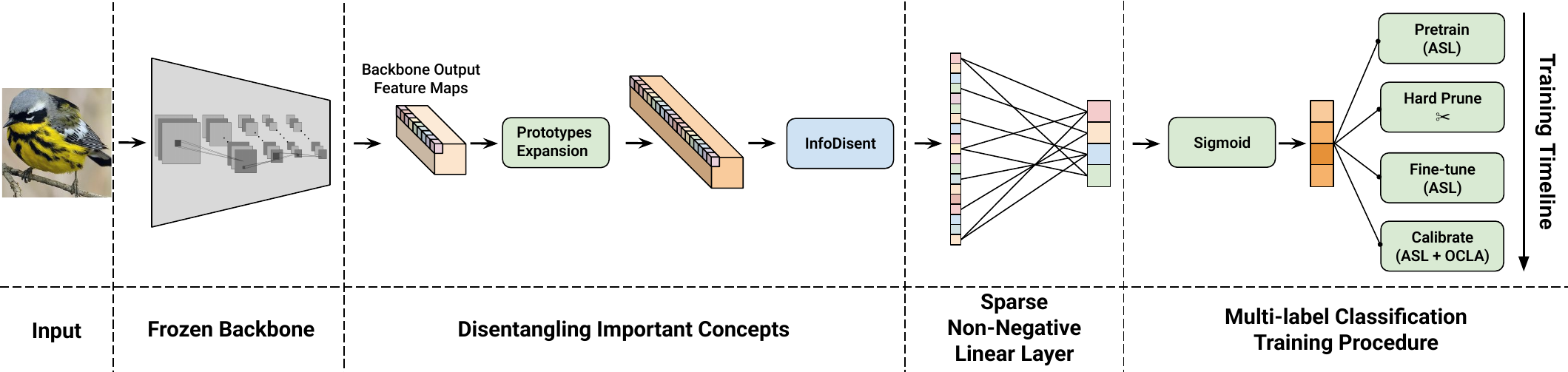}
\caption{Main components of the SIDE interpretability method.}
\label{fig:side_components}
\end{figure*}

This section presents SIDE (Figure~\ref{fig:side_components}), which builds on InfoDisent through 
three architectural improvements—Prototype Expansion, Sigmoid activations with ASL, a Sparse Scores Sheet—and a structured training procedure. 

Training proceeds in four stages: pretraining \(\rightarrow\) hard pruning \(\rightarrow\) fine-tuning \(\rightarrow\) calibration, with the final stage introducing OCLA-based regularization (Section \ref{sec:ocla}).

\subsection{Prototypes Expansion}

A key limitation of InfoDisent on large‐scale datasets (e.g.\ ImageNet‐1K/22K) is that the number of prototypes is tied directly to the channel dimension \(C\) of the backbone’s feature map. Modern transformers use moderate channel sizes—e.g.\ ViT-Base \cite{dosovitskiy2021imageworth16x16words}, and SwinV2-Small \cite{liu2022swintransformerv2scaling} each have \(C=768\), while ViT-Large and SwinV2-Base, use \(C=1024\). Even on ImageNet-1K (1,000 classes), this yields around one prototype per class; on ImageNet-22K the ratio drops to fewer than one prototype for every 22 classes, severely constraining model expressivity.

SIDE addresses this by decoupling the prototype count from the backbone’s native channels via Prototype Expansion. We apply a trainable \(1\times1\) convolutional layer to the frozen backbone output, projecting its \(C\)-dimensional features up to a higher dimension \(C'\). Because this expansion is channel‐only, it preserves the spatial localization and patch‐activation patterns of the original feature map. 

Moreover, the choice of \(C'\) is not critical to the final effectiveness or interpretability of SIDE. Thanks to its sparsity mechanisms, SIDE can prune a large number of redundant connections. Thus, selecting \(C' \gg C\) allows us to define an ample upper bound on the number of prototypes without loss in compactness or interpretability (see Figure~\ref{fig:weights_over_epochs}).

\subsection{Multilabel Classification}

\textbf{Sigmoid versus Softmax.}  
Many prototypical models—including InfoDisent—operate in a quasi-multilabel setting (i.e.\ a single prototype can support multiple classes). To capture this, SIDE replaces the conventional Softmax activation with independent Sigmoid functions on the unnormalized class logits. This change lets each class attain a high similarity score without suppressing others, more faithfully reflecting the overlaps in prototypical space.

Natural loss function choice for Sigmoid is Binary Cross-Entropy (BCE). BCE treats each class prediction independently, preserving informative gradients from both positive and negative labels. By contrast, CE + Softmax introduces relative comparisons between class scores, as Softmax is invariant to adding any constant to the logits, thus coupling every class score to all the others. This coupling can mask true ambiguity: a class may appear highly confident solely because its competitors scored low.

\begin{figure}[t!] 
\centering
\includegraphics[width=\columnwidth]{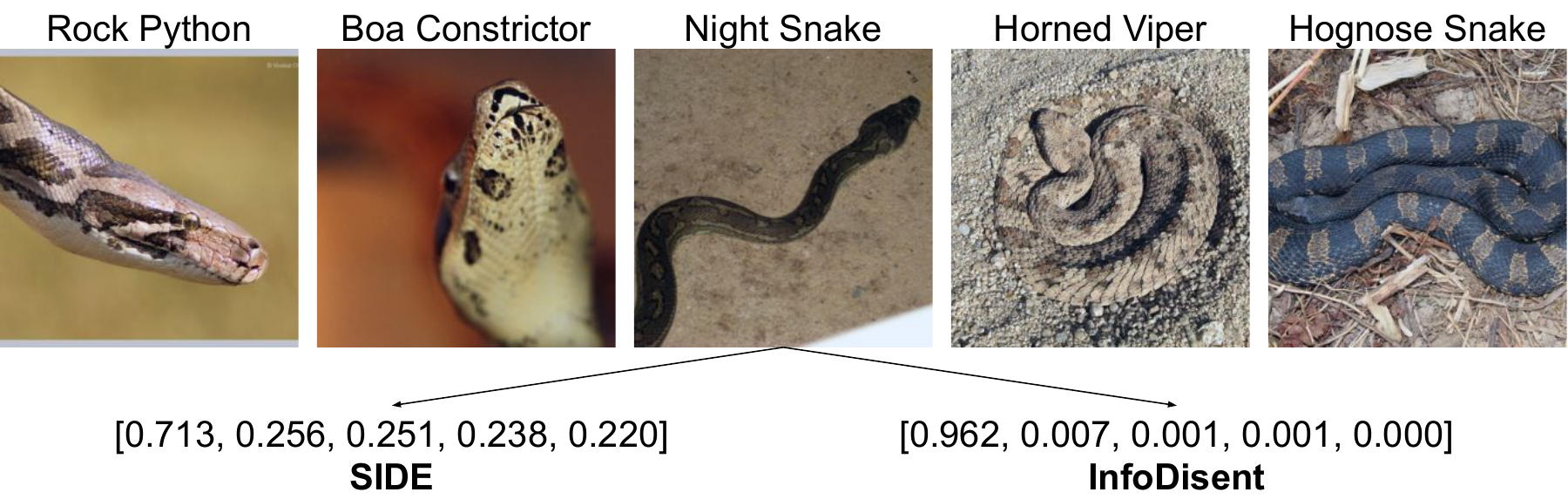}
\caption{Class probability distributions for an ImageNet Rock Python: SIDE’s Sigmoid classifier (left) assigns substantial scores to several visually and semantically similar snake subclasses, whereas InfoDisent’s Softmax classifier (right) concentrates nearly all mass on its top prediction.}
\label{fig:compare_proba}
\vspace{-1.6em}
\end{figure}

In addition to it, Softmax classifiers are often poorly calibrated \cite{guo2017calibrationmodernneuralnetworks} and tend to assign nearly all probability mass to a single class—even when multiple visually or semantically similar classes are present—thereby masking model uncertainty. In contrast, SIDE’s independent Sigmoid activations mitigate this overconfidence: as Figure \ref{fig:compare_proba} shows, Sigmoid probabilities are distributed across the top-5 snake subclasses, more faithfully reflecting their shared features and enhancing interpretability.

\textbf{Loss function choice.}
Despite these advantages, BCE introduces its own challenges, particularly in the presence of class imbalance and a growing number of labels. In such settings, the cumulative signal from numerous negative classes can dominate the gradients, overwhelming the positive signal. This issue is further amplified when backpropagating through the sparse, non-negative Scores Sheet of SIDE (§\ref{sec:training_procedure}), where abrupt shifts in prototype activation magnitudes can destabilize training or trigger unintended pruning of many prototype weights in a single update.

More nuanced control of signal from negative samples, especially during different stages of training (pretraining and fine-tuning §\ref{sec:training_procedure}) motivated us to look at ASL. As we show through our experiments it allows for stable training with pruning of weights in sparse Scores Sheet, has faster convergence than Cross-Entropy and achieves better performance on accuracy metrics during pretraining.

\textbf{One Correct Label Activation.}
\label{sec:ocla}
Let \( t \in (0, 1) \) be an activation threshold and \( p_i \in (0, 1) \) the predicted probability for class \( i \in \{1, \ldots, C\} \), where \( C \) is the number of classes. We define a class as activated for a given sample if \( p_i > t \).

A key interpretability benefit of replacing the Softmax with Sigmoid activation is that class probability scores become independent and globally meaningful, rather than relative. This enables the model to express uncertainty more explicitly: it may assign high confidence to multiple classes, or low scores to all classes, abstaining from a decision.

This behavior is useful in concept-based models. When multiple classes are activated, the corresponding prototypes can be inspected to understand similarities between these classes. Conversely, when no class is activated, the absence of strong evidence may also be interpretable. In our \textbf{SIDE}, we define the local explanation for a sample as the set of prototypes associated with all activated classes. When no labels are activated we include set of prototypes of the maximally activated label as the reason behind model abstaining from the decision has to be considered as well.

To evaluate whether SIDE produces confident and interpretable predictions on one-class classification tasks, we introduce the One Correct Label Activation (OCLA) metric. This metric computes the proportion of test samples for which exactly one class is activated. Let \( N \) be the number of test samples, \( Y \in \{1, \ldots, C\}^N \) the ground-truth labels, and \( P \in [0, 1]^{N \times C} \) the matrix of predicted class probabilities. Then OCLA is defined as:

\begin{equation}
\text{OCLA}(P, Y) = \frac{1}{N} \sum_{i=1}^{N} \mathbbm{1} \left[ \sum_{j=1}^{C} \mathbbm{1}(P_{ij} > t)  \ \& \  j=Y_i \right]
\label{eq:ocla_metric}
\end{equation}

This metric favors when model produces confident predictions for a single class, which simplifies interpretation by limiting the number of prototypes to be examined. Higher OCLA values indicate sharper, more selective predictions.

We use a default threshold of \( t = 0.5 \) and find that the ASL naturally promotes high OCLA scores. However, following hard prototype pruning, the model often requires additional recalibration to encourage confident, single-label predictions. That is why, we introduce OCLA loss — a regularization applied at the final stage of training to ensure that exactly one class probability exceeds threshold \( t \).

\def\relu{\mathrm{ReLU}}

\begin{align}
\text{OCLA}(P, Y, t) 
&= \frac{1}{N} \sum_{i=1}^{N} \bigg( 
\sum_{\substack{j = 1 \\ j \neq Y_i}}^{C} \relu(P_{ij} - t) \nonumber \\
&\phantom{=} 
+ s \cdot \relu(P_{i, Y_i} - t) 
- s \cdot (P_{i, Y_i} - t)
\bigg)
\label{eq:ocla_loss}
\end{align}

The OCLA loss addresses three distinct cases:
\begin{enumerate}
    \item \textbf{At least one incorrect class is activated, but the correct class is not.} In this case, all incorrectly activated classes (\( P_{ij} > t, j \neq Y_i \)) are penalized, while the correct class \( P_{i, Y_i} \) is encouraged to exceed the threshold via the final term.
    \item \textbf{Multiple classes are activated, including the correct one.} Here, only the incorrect active classes are penalized, and the correct activation is left unaffected due to cancellation between the second and third terms.
    \item \textbf{No class is activated.} The model is encouraged to push the correct class above the threshold by a negative loss term proportional to \( -s \cdot (P_{i, Y_i} - t) \).
\end{enumerate}

The scalar \( s \in \mathbb{R} \) controls the strength of the corrective signal applied to the correct class relative to the penalization of incorrect activations. 


\subsection{Training Procedure \& Pruning}
\label{sec:training_procedure}

As discussed in \ref{sec:info-disent}, InfoDisent enforces positive contributions from each prototype by applying an element-wise absolute value to the classification weights (Eq. \ref{eq:infodisent_ll}) While this constraint supports interpretability through additivity, it does not enforce sparsity—small weights remain trainable and can grow over time, potentially diluting the explanatory power of individual prototypes.

In contrast, \textbf{SIDE} encourages sparsity by replacing the absolute-value operation with ReLU, applying Sigmoid activations in place of Softmax and optimizing the resulting probabilities using ASL:
\begin{equation}
p = \sigma(\max(W, 0) \cdot v).
\label{eq:side_linear}
\end{equation}
where \( \sigma(\cdot) \) denotes the sigmoid function.
\( \max(W, 0) \) ensures that once a weight is zeroed, it remains inactive. To bias learning toward activating all prototypes initially, we initialize the weights as \( W \sim \mathcal{N}(1.0, 0.1) \).

The training of SIDE is structured into four stages: Pretraining, Hard pruning, Fine-tuning and Calibration

\textbf{Pretraining.} The model is trained using the modified Scores Sheet and ASL. During this stage, a relatively high learning rate and small values of (\( \large \gamma_{-} \), \( \large p_m \)) ASL parameters encourage the model to naturally down-weight and prune uninformative connections.

\textbf{Hard Pruning.} After pretraining, we apply hard pruning by leaving only top-\( k \) scores of \( W \), where \( k = A * C \) and \( A \) is desired average number of prototypes activated per class and zeroing-out all the other weights.

\textbf{Fine-tuning.} Next, the model is fine-tuned with a lower learning rate and higher loss parameters values to adapt to the pruned set of connections and recover predictive performance, while potentially identifying additional redundant connections for further pruning.

\textbf{Calibration.} At the final stage of training, the OCLA regularization term is added to ASL to further increase the separation between the correct class probability and those of all other classes, improving OCLA as well as Local Size of the model. We denote OCLA term coefficient during Calibration stage as \( \lambda \).

\section{Experiments}

\begin{figure*}[t!] 
\centering
\includegraphics[width=0.8\textwidth]{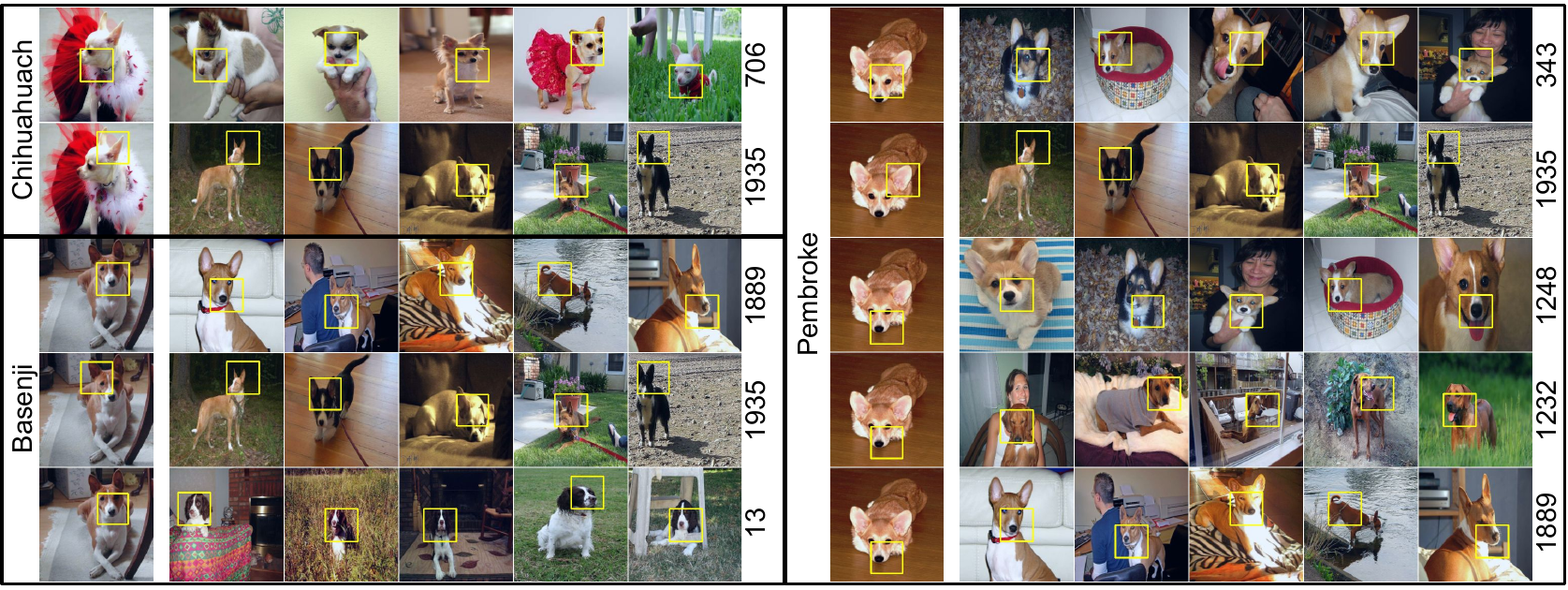}
\caption{SIDE explanation for three Stanford Dogs classes using ResNet-50. With a hard pruning threshold of \( A = 5 \), SIDE learns to share semantically meaningful prototypes across classes. A prototype capturing large, pointy ears (1935) is reused by Chihuahua, Basenji, and Pembroke, while an oblong face with a black nose (1889) is shared between Basenji and Pembroke.
SIDE yields compact explanations with 2, 3, and 5 prototypes for Chihuahua, Basenji, and Pembroke, respectively.}
\label{fig:main_dogs}
\vspace{-2em}
\end{figure*}

SIDE is evaluated on the well-established benchmarks in literature of concept-based models: CUB-200-2011 \cite{wah_2023_cvm3y-5hh21} (200 bird species), Stanford Cars \cite{6755945} (196 car models) and Stanford Dogs \cite{KhoslaYaoJayadevaprakashFeiFei_FGVC2011} (120 dogs species) and compared to current state-of-the-art methods, including InfoDisent. Additionally, we validate our approach on ImageNet \cite{russakovsky2015imagenetlargescalevisual}. For fine-grained datasets SIDE is evaluated on common convolutional backbones Resnet-50 \cite{he2015deepresiduallearningimage} and Densnet-121 \cite{huang2018denselyconnectedconvolutionalnetworks}, whereas on ImageNet SIDE is applied to SwinV2-S. We provide more details on experiments setup in Supplementary Materials and we make code available.

\subsection{Fine-grained Datasets}

As shown in Table~\ref{tab:merged_comparison}, SIDE achieves a significantly more compact explanation structure than InfoDisent, with reductions of over \(10\times\) in Global Size and more than \(300\times\) in Local Size, while maintaining comparable top-1 accuracy across all datasets and backbones.

On both ResNet-50 and DenseNet-121, SIDE preserves classification performance—deviating by less than \(0.7\%\) from InfoDisent on any dataset—while producing explanations that are orders of magnitude sparser. For instance, on Stanford Dogs with ResNet-50, SIDE achieves \(86.8\%\) accuracy using just \(6.5\) prototypes per image, compared to \(2048\) used by InfoDisent. Among all methods evaluated SIDE underperforms only against TesNet and ST-ProtoPNet on CUB when applied to DenseNet-121, however SIDE's backbone on CUB was pretrained on ImageNet and not on iNaturalist \cite{vanhorn2018inaturalistspeciesclassificationdetection}, which usually provides a substantial increase in accuracy. Compared to other interpretable models with similar prototype budgets (e.g., ProtoTree, PIP-Net), SIDE consistently delivers higher accuracy while maintaining or exceeding sparsity. 

\vspace{-0.3em}

\begin{table*}[t]
\centering
\small
\caption{Top-1 accuracy (Acc), total number of active prototypes (Global Size), and number of prototypes used to explain a single prediction (Local Size) on cropped CUB-200-2011, cropped Stanford Cars, and full Stanford Dogs datasets. SIDE achieves comparable accuracy to InfoDisent, while using significantly fewer prototypes globally and locally. It also outperforms other methods with similar Local Size, demonstrating a favorable trade-off between interpretability and predictive performance.}
\label{tab:merged_comparison}
\begin{tabular}{l l 
  *{3}{ccc}}
\toprule
 & \textbf{Method}
   & \multicolumn{3}{c}{\textbf{CUB‐200‐2011}}
   & \multicolumn{3}{c}{\textbf{Stanford Cars}}
   & \multicolumn{3}{c}{\textbf{Stanford Dogs}} \\
\cmidrule(lr){3-5} \cmidrule(lr){6-8} \cmidrule(lr){9-11}
 & 
   & \textbf{Acc} & \textbf{Global Size} & \textbf{Local Size}
   & \textbf{Acc} & \textbf{Global Size} & \textbf{Local Size}
   & \textbf{Acc} & \textbf{Global Size} & \textbf{Local Size} \\
\midrule
\multirow{9}{*}{\rotatebox[origin=c]{90}{\textbf{ResNet‐50}}}
  & Baseline            & 83.2 & –    & – & 93.1 & –    & – & 87.4 & –    & – \\
  & InfoDisent          & 83.0 & 2048 & 2048 & 92.9 & 2048 & 2048 & 86.6 & 2048 & 2048 \\
& \textbf{SIDE (Ours)}& 82.7 & 310  & 6.3 & 92.8 & 360  & 7.9 & 86.8 & 238  & 6.5 \\
\cdashline{2-11}
\addlinespace[2pt]
  & ProtoPool           & – & –  & – & 88.9 & 195  & 195 & –    & –    & – \\
  & ProtoTree           & 82.2 & 202  & 8.3 & 86.6 & 195  & 8.5 & –    & –    & – \\
  & PIP‐Net             & 82.0 & 731  & 12 & 86.5 & 669  & 11 & –    & –    & – \\
  & ProtoPNet           & 79.2    & 2000 & 2000   & 86.1    & 1960    & 1960   & 76.4 & 1200 & 1200 \\
  & ST‐ProtoPNet        & –    & –    & –   & –    & –    & –   & 84.0 & 1200 & 1200 \\
  & TesNet              & –    & –    & –   & –    & –    & –   & 82.4 & 1200 & 1200 \\
\midrule
\multirow{8}{*}{\rotatebox[origin=c]{90}{\textbf{DenseNet‐121}}}
  & Baseline            & 81.8 & –    & – & 92.1 & –    & – & 84.1 & –    & – \\
  & InfoDisent          & 82.6 & 1024 & 1024 & 92.7 & 1024 & 1024 & 83.8 & 1024 & 1024 \\
  & \textbf{SIDE (Ours)}& 82.0 & 266  & 5.5 & 92.3 & 444  & 4.8 & 83.1 & 163  & 7.27 \\
\cdashline{2-11}
\addlinespace[2pt]
  & ProtoPNet           & 79.2 & 2000 & 2000 & 86.8 & 1960 & 1960 & 72.0 & 1200 & 1200 \\
  & ProtoPShare         & 74.7 & 400  & 400 & 84.8 & 480  & 480 & –    & –    & – \\
  & ProtoPool           & 73.6 & 202  & 202 & 86.4 & 195  & 195 & –    & –    & – \\
  & ST‐ProtoPNet        & 85.4 & 2000 & 2000 & 92.3 & 1960 & 1960 & 79.4 & 1200 & 1200 \\
  & TesNet              & 84.8 & 2000 & 2000 & 92.0 & 1960 & 1960 & 80.3 & 1200 & 1200 \\
\bottomrule
\end{tabular}
\end{table*}

\subsection{Scaling to ImageNet}

Table~\ref{tab:imagenet_results} demonstrates that SIDE successfully extends prototypical interpretability to large-scale recognition tasks such as ImageNet—an area where most existing concept-based models fail to operate effectively.

In its default configuration (A=10), SIDE achieves \(82.8\%\) top-1 accuracy—just \(0.6\,\text{pp}\) below the SwinV2-S baseline—while reducing the average Local Size to only \(8.6\) prototypes. When sparsity is pushed further (A=7), SIDE still matches InfoDisent’s accuracy (\(81.6\%\) vs.\ \(81.4\%\)) with just \(6.9\) prototypes per prediction.

While Prototype Expansion increases the size of the global prototype pool (\(947\text{--}1340\) vs.\ \(768\) for InfoDisent), SIDE’s pruning ensures that each decision is supported by only the most informative concepts. This results in explanations that are over \(100\times\) smaller than those of InfoDisent, while closing \(75\%\) of its accuracy gap to the backbone. 

\begin{table}[t]
\centering
\small
\caption{Top-1 ImageNet accuracy, total number of active prototypes (Global Size), and number of prototypes used to explain a single prediction (Local Size) on ImageNet. Compared to InfoDisent on a SwinV2-S backbone, SIDE achieves higher accuracy while consistently using fewer prototypes per prediction, even when the Global Size is larger.}
\label{tab:imagenet_results}
\begin{tabular}{l ccc}
\toprule
\textbf{Model}
  & \multicolumn{3}{c}{\textbf{ImageNet}} \\
\cmidrule(lr){2-4}
               & Accuracy  & Global Size  & Local Size  \\
\midrule
Baseline     & 83.4     & --        & --       \\
InfoDisent          & 81.4     & 768      & 768     \\
\textbf{SIDE} (A=10) & 82.6 & 1340 & 8.6  \\
\textbf{SIDE} (A=7) & 81.6 & 947 & 6.9  \\
\bottomrule
\end{tabular}
\end{table}

\subsection{Interpretability}
\begin{figure*}[t!] 
\centering
\includegraphics[width=0.9\textwidth]{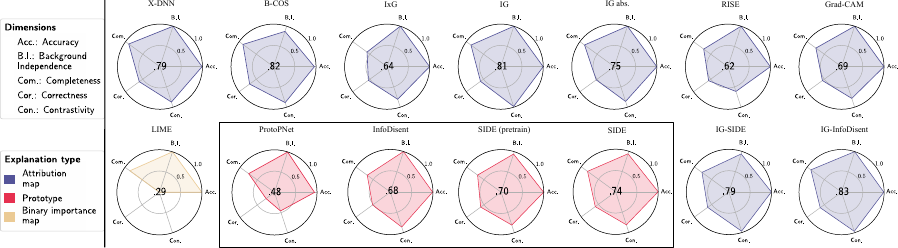}
\caption{Results of various interpretability methods on Funny Birds framework with Resnet-50 as backbone. SIDE (pretrain) was evaluated after the pretraining stage, SIDE is a fully trained version (A=10). SIDE achieves the best results among all Prototype-based methods.}
\label{fig:funny}
\vspace{-1.5em}
\end{figure*}

To quantitatively assess the interpretability of SIDE, we evaluate it on the FunnyBirds~\cite{hesse2023funnybirds} benchmark—a dedicated XAI evaluation framework built around synthetic, part-based image classification. FunnyBirds enables fine-grained, multi-dimensional comparison of explanation quality by providing part annotations and controlled feature manipulations. We focus on three core dimensions: correctness, which measures how faithfully the explanation reflects the model’s decision-making process (via deletion tests); completeness, which quantifies whether the explanation covers all critical decision evidence; and contrastivity, which reflects how well the explanation distinguishes the predicted class from alternatives. Together, these dimensions provide a comprehensive picture of explanation alignment with the underlying model behavior.

Prior concept-based methods like ProtoPNet underperform on FunnyBirds, achieving particularly low scores on correctness and contrastivity. InfoDisent significantly improves over this baseline by disentangling feature channels and encouraging part-level alignment, yet it still struggles along correctness and completness dimensions.

SIDE outperforms both ProtoPNet and InfoDisent across completness and correctness dimensions while preserving same contrastivity level. Even in its pre-trained form—before pruning is applied—SIDE surpasses InfoDisent in overall explanation quality (1578 Global Size vs 2048, \( 13K \) Scores Sheet active connections vs \( 102K \)), indicating that its sparse, disentangled prototype space already aligns more tightly with model behavior. 

The full pruned version of SIDE (64 Global Size, 500 connections) further improves completeness by removing weak or redundant prototype–class connections and concentrating the decision explanation into a compact, non-overlapping subset of highly informative parts. This refinement causes only a negligible drop in background independence, suggesting SIDE’s sparsity does not lead to overfitting on foreground or artifacts.

For completeness, we also report the performance of hybrid methods that combine Integrated Gradients \cite{sundararajan2017axiomaticattributiondeepnetworks} with prototype-based explanations, namely IG-InfoDisent and IG-SIDE. We note that IG-InfoDisent outperforms IG-SIDE. However, since these methods blend fundamentally different explanation mechanisms, we do not consider them primary indicators of prototypical explanation quality and include them only for reference.

\subsection{Ablation Studies}
We conduct a series of ablations to disentangle the contributions of SIDE’s components to interpretability and performance. In particular, we analyze the dynamics of prototype pruning, the effect of post-pruning fine-tuning, the need for a separate calibration stage, and the structure of learned prototype weights.

\begin{figure}[t]
\centering
\includegraphics[width=0.8\columnwidth]{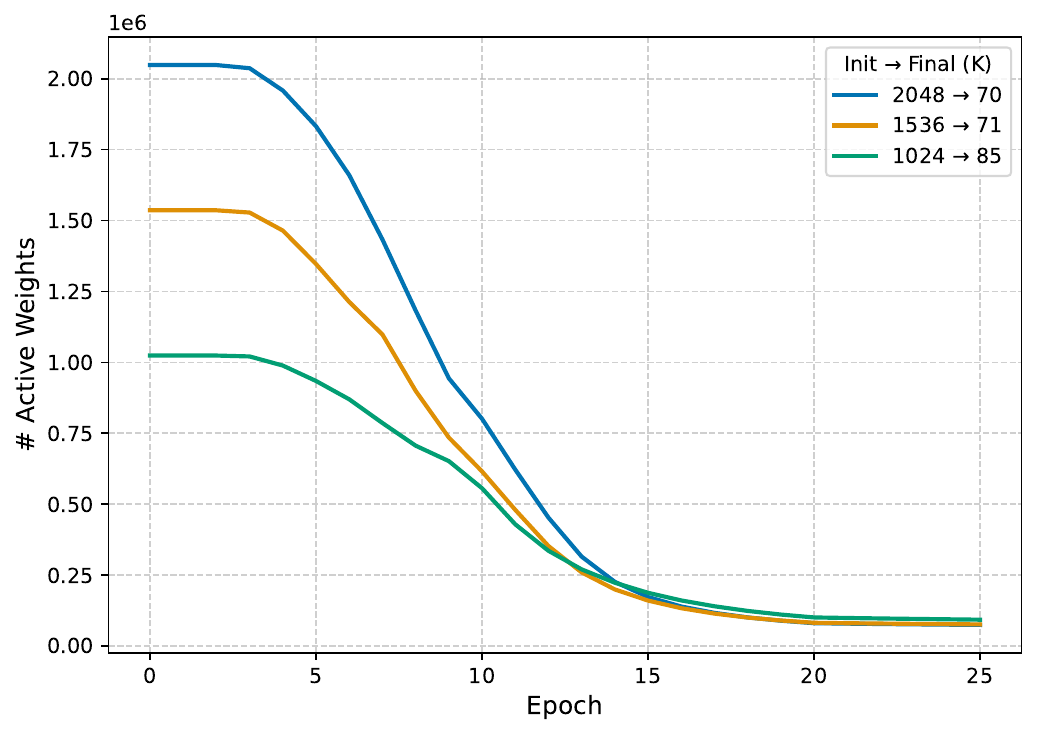}
\caption{Number of active weights in the Scores Sheet of SIDE during pretraining on ImageNet-1K with a SwinV2-S backbone. Only the first 25 of 50 epochs are shown for clarity. Despite different Prototype Expansion dimensions (\(C' = 2048, 1536, 1024\)), all models converge to a similar number of active weights after approximately 20 epochs.}
\label{fig:weights_over_epochs}
\end{figure}

\begin{figure}[t]
\centering
\includegraphics[width=0.95\columnwidth]{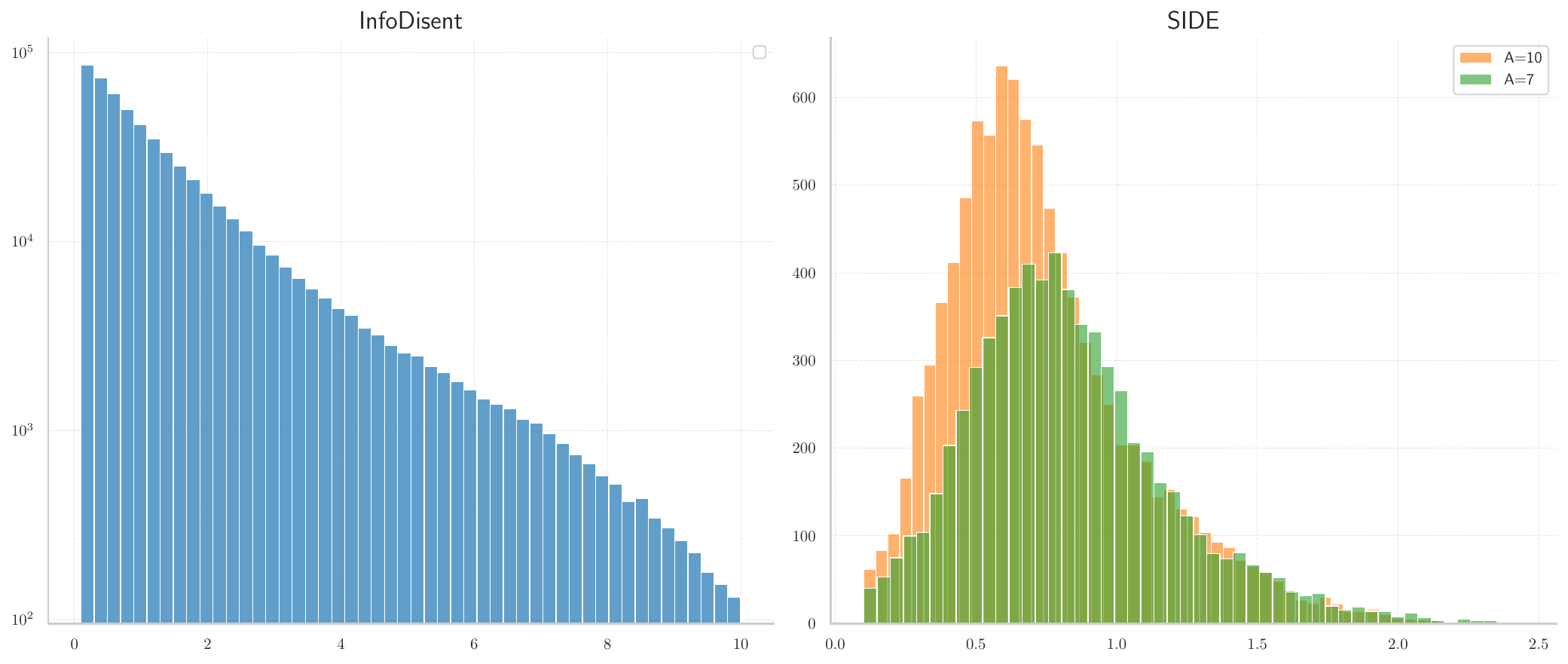}
\caption{Distributions of Scores Sheet active weights for InfoDisent and SIDE applied to SwinV2-S on ImageNet-1K. The left plot shows InfoDisent, with weights plotted on a logarithmic y-axis. The right plot shows SIDE in two variants with hard pruning thresholds $A{=}10$ and $A{=}7$. Compared to InfoDisent, SIDE produces significantly sparser and more concentrated weight distributions with sharply peaked modes and limited tails.}
\label{fig:weights_comparison}
\vspace{-2em}
\end{figure}

\textbf{Sparse weight emergence during pretraining.}
Figure~\ref{fig:weights_over_epochs} shows how the pruning effect of ASL emerges organically during pretraining. Despite using no explicit sparsity loss and varying the Prototype Expansion dimensionality \( C' \in \{1024,1536,2048\} \), all models converge to a similar number of active prototype–class connections after around 20 epochs. This indicates that the ASL, when applied with a non-negative ReLU-linear head, effectively prunes uninformative weights and that the final number of used connections is not tightly tied to the expansion dimension \( C'\).

\textbf{Effect of hard pruning and importance of fine-tuning.}
As shown in Supplementary Section \( 2.1 \), hard pruning alone causes a steep accuracy drop, especially for aggressive thresholds. While pruning significantly reduces the Local Size and Global Size, the model's accuracy post-pruning degrades without subsequent fine-tuning. However, SIDE reliably recovers performance during fine-tuning across a wide range of pruning thresholds. The gap between post-prune and final accuracy emphasizes that pruning must be followed by adaptation to the pruned structure.

\textbf{Impact of calibration stage on interpretability.}
Hard pruning not only reduces prototype number but also causes a notable drop in OCLA score (see Supplementary Materials Section \( 2.2 \)). This drop indicates that SIDE becomes under-calibrated, i.e. the model’s predictions are less often driven by a single class and explanation are less focused. Importantly, fine-tuning without any additional objective (\( \lambda=0\)) fails to fully recover the OCLA score.

To address this, we introduce a final calibration stage using the OCLA regularization. As shown in Supplementary Section \( 2.2 \), SIDE models trained with \( \lambda \in \{100,200\} \) achieve consistently higher OCLA scores and smaller Local Size, with \( \lambda=200 \) yielding the best trade-off.

\textbf{Concentration of prototype weights.}
Finally, in Figure~\ref{fig:weights_comparison}, we visualize the distribution of active prototype-class connections in the Scores Sheet for InfoDisent and SIDE. While InfoDisent spreads importance across many small weights and values of active connections vary by a magnitude, SIDE produces sharply peaked, concentrated distributions, especially after pruning. This suggests that nearly all retained prototypes contribute meaningfully to the decision, resulting in compact and faithful explanations.

\section{Conclusions}

In this work, we introduced SIDE, a novel method designed to generate sparse and compact explanations for prototypical-parts-based models. SIDE not only maintains high performance across both fine-grained datasets and large-scale computer vision benchmarks like ImageNet but also offers more concise and user-friendly explanations. 

\textbf{Limitations and Future Work.}
The primary limitation of SIDE, shared by many prototypical-parts models, is its complex multi-stage training procedure. Additionally, SIDE inherits certain limitations from the InfoDisent approach. In future research, we aim to explore self-supervised learning of concept-based image representations to reduce the reliance on extensive supervision during the learning process.

\textbf{Impact.}
This work highlights the importance of providing concise and sparse explanations to users of AI systems. We underscore that model can generate misleading explanations, making it crucial to check its correctness to prevent misinformation.

\section*{Acknowledgements}

This research was partially funded by the National Science Centre, Poland, grants no. 2020/39/D/ST6/01332 (work by Łukasz Struski), and 2023/49/B/ST6/01137 (work by Viktar Dubovik, Jacek Tabor, and Dawid Rymarczyk). Some experiments were performed on servers purchased with funds from the flagship project entitled “Artificial Intelligence Computing Center Core Facility” from the DigiWorld Priority Research Area within the Excellence Initiative – Research University program at Jagiellonian University in Kraków.
{
    \small
    \bibliographystyle{ieeenat_fullname}
    \bibliography{mybibfile}

\begin{thebibliography}{50}
\providecommand{\natexlab}[1]{#1}
\providecommand{\url}[1]{\texttt{#1}}
\expandafter\ifx\csname urlstyle\endcsname\relax
  \providecommand{\doi}[1]{doi: #1}\else
  \providecommand{\doi}{doi: \begingroup \urlstyle{rm}\Url}\fi

\bibitem[Afnan et~al.(2021)Afnan, Liu, Conitzer, Rudin, Mishra, Savulescu, and
  Afnan]{afnan2021interpretable}
Michael Anis~Mihdi Afnan, Yanhe Liu, Vincent Conitzer, Cynthia Rudin, Abhishek
  Mishra, Julian Savulescu, and Masoud Afnan.
\newblock Interpretable, not black-box, artificial intelligence should be used
  for embryo selection.
\newblock \emph{Human Reproduction Open}, 2021.

\bibitem[Alvarez~Melis and Jaakkola(2018)]{NEURIPS2018_3e9f0fc9}
David Alvarez~Melis and Tommi Jaakkola.
\newblock Towards robust interpretability with self-explaining neural networks.
\newblock In \emph{Advances in Neural Information Processing Systems}. Curran
  Associates, Inc., 2018.

\bibitem[Barnett et~al.(2021)Barnett, Schwartz, Tao, Chen, Ren, Lo, and
  Rudin]{barnett2021case}
Alina~Jade Barnett, Fides~Regina Schwartz, Chaofan Tao, Chaofan Chen, Yinhao
  Ren, Joseph~Y Lo, and Cynthia Rudin.
\newblock A case-based interpretable deep learning model for classification of
  mass lesions in digital mammography.
\newblock \emph{Nature Machine Intelligence}, 3\penalty0 (12):\penalty0
  1061--1070, 2021.

\bibitem[Ben-Baruch et~al.(2021)Ben-Baruch, Ridnik, Zamir, Noy, Friedman,
  Protter, and
  Zelnik-Manor]{benbaruch2021asymmetriclossmultilabelclassification}
Emanuel Ben-Baruch, Tal Ridnik, Nadav Zamir, Asaf Noy, Itamar Friedman, Matan
  Protter, and Lihi Zelnik-Manor.
\newblock Asymmetric loss for multi-label classification, 2021.

\bibitem[Bibal et~al.(2020)Bibal, Lognoul, de~Streel, and Frénay]{Bibal2020}
Adrien Bibal, Michael Lognoul, Alexandre de Streel, and Benoît Frénay.
\newblock Legal requirements on explainability in machine learning.
\newblock \emph{Artificial Intelligence and Law}, 29\penalty0 (2):\penalty0
  149–169, 2020.

\bibitem[B{\"o}hle et~al.(2022)B{\"o}hle, Fritz, and Schiele]{bohle2022b}
Moritz B{\"o}hle, Mario Fritz, and Bernt Schiele.
\newblock B-cos networks: Alignment is all we need for interpretability.
\newblock In \emph{Proceedings of the IEEE/CVF Conference on Computer Vision
  and Pattern Recognition}, pages 10329--10338, 2022.

\bibitem[Borycki et~al.(2025)Borycki, Tr{\k{e}}dowicz, Janusz, Tabor, Spurek,
  Lewicki, and Struski]{borycki2025epic}
Piotr Borycki, Magdalena Tr{\k{e}}dowicz, Szymon Janusz, Jacek Tabor,
  Przemys{\l}aw Spurek, Arkadiusz Lewicki, and {\L}ukasz Struski.
\newblock Epic: Explanation of pretrained image classification networks via
  prototype.
\newblock \emph{arXiv preprint arXiv:2505.12897}, 2025.

\bibitem[Brendel and Bethge(2019)]{brendel2018approximating}
Wieland Brendel and Matthias Bethge.
\newblock Approximating {CNN}s with bag-of-local-features models works
  surprisingly well on imagenet.
\newblock In \emph{International Conference on Learning Representations}, 2019.

\bibitem[Buhrmester et~al.(2019)Buhrmester, Münch, and
  Arens]{buhrmester2019analysisexplainersblackbox}
Vanessa Buhrmester, David Münch, and Michael Arens.
\newblock Analysis of explainers of black box deep neural networks for computer
  vision: A survey, 2019.

\bibitem[Chen et~al.(2019)Chen, Li, Tao, Barnett, Su, and
  Rudin]{chen2019lookslikethatdeep}
Chaofan Chen, Oscar Li, Chaofan Tao, Alina~Jade Barnett, Jonathan Su, and
  Cynthia Rudin.
\newblock This looks like that: Deep learning for interpretable image
  recognition, 2019.

\bibitem[Donnelly et~al.(2022)Donnelly, Barnett, and
  Chen]{donnelly2022deformable}
Jon Donnelly, Alina~Jade Barnett, and Chaofan Chen.
\newblock Deformable protopnet: An interpretable image classifier using
  deformable prototypes.
\newblock In \emph{Proceedings of the IEEE/CVF conference on computer vision
  and pattern recognition}, pages 10265--10275, 2022.

\bibitem[Dosovitskiy et~al.(2021)Dosovitskiy, Beyer, Kolesnikov, Weissenborn,
  Zhai, Unterthiner, Dehghani, Minderer, Heigold, Gelly, Uszkoreit, and
  Houlsby]{dosovitskiy2021imageworth16x16words}
Alexey Dosovitskiy, Lucas Beyer, Alexander Kolesnikov, Dirk Weissenborn,
  Xiaohua Zhai, Thomas Unterthiner, Mostafa Dehghani, Matthias Minderer, Georg
  Heigold, Sylvain Gelly, Jakob Uszkoreit, and Neil Houlsby.
\newblock An image is worth 16x16 words: Transformers for image recognition at
  scale, 2021.

\bibitem[Gairola et~al.(2025)Gairola, B{\"o}hle, Locatello, and
  Schiele]{gairola2025probe}
Siddhartha Gairola, Moritz B{\"o}hle, Francesco Locatello, and Bernt Schiele.
\newblock How to probe: Simple yet effective techniques for improving post-hoc
  explanations.
\newblock \emph{ICLR}, 2025.

\bibitem[Gee et~al.(2019)Gee, Garcia-Olano, Ghosh, and
  Paydarfar]{gee2019explaining}
Alan~H Gee, Diego Garcia-Olano, Joydeep Ghosh, and David Paydarfar.
\newblock Explaining deep classification of time-series data with learned
  prototypes.
\newblock In \emph{CEUR workshop proceedings}, page~15. NIH Public Access,
  2019.

\bibitem[Guo et~al.(2017)Guo, Pleiss, Sun, and
  Weinberger]{guo2017calibrationmodernneuralnetworks}
Chuan Guo, Geoff Pleiss, Yu Sun, and Kilian~Q. Weinberger.
\newblock On calibration of modern neural networks, 2017.

\bibitem[Hase et~al.(2019)Hase, Chen, Li, and Rudin]{hase2019interpretable}
Peter Hase, Chaofan Chen, Oscar Li, and Cynthia Rudin.
\newblock Interpretable image recognition with hierarchical prototypes.
\newblock In \emph{Proceedings of the AAAI Conference on Human Computation and
  Crowdsourcing}, pages 32--40, 2019.

\bibitem[He et~al.(2015)He, Zhang, Ren, and
  Sun]{he2015deepresiduallearningimage}
Kaiming He, Xiangyu Zhang, Shaoqing Ren, and Jian Sun.
\newblock Deep residual learning for image recognition, 2015.

\bibitem[Hesse et~al.(2023)Hesse, Schaub-Meyer, and Roth]{hesse2023funnybirds}
Robin Hesse, Simone Schaub-Meyer, and Stefan Roth.
\newblock Funnybirds: A synthetic vision dataset for a part-based analysis of
  explainable ai methods.
\newblock In \emph{Proceedings of the IEEE/CVF International Conference on
  Computer Vision}, pages 3981--3991, 2023.

\bibitem[Horn et~al.(2018)Horn, Aodha, Song, Cui, Sun, Shepard, Adam, Perona,
  and Belongie]{vanhorn2018inaturalistspeciesclassificationdetection}
Grant~Van Horn, Oisin~Mac Aodha, Yang Song, Yin Cui, Chen Sun, Alex Shepard,
  Hartwig Adam, Pietro Perona, and Serge Belongie.
\newblock The inaturalist species classification and detection dataset, 2018.

\bibitem[Huang et~al.(2018)Huang, Liu, van~der Maaten, and
  Weinberger]{huang2018denselyconnectedconvolutionalnetworks}
Gao Huang, Zhuang Liu, Laurens van~der Maaten, and Kilian~Q. Weinberger.
\newblock Densely connected convolutional networks, 2018.

\bibitem[Keswani et~al.(2022)Keswani, Ramakrishnan, Reddy, and
  Balasubramanian]{keswani2022proto2proto}
Monish Keswani, Sriranjani Ramakrishnan, Nishant Reddy, and Vineeth~N
  Balasubramanian.
\newblock Proto2proto: Can you recognize the car, the way i do?
\newblock In \emph{Proceedings of the IEEE/CVF Conference on Computer Vision
  and Pattern Recognition}, pages 10233--10243, 2022.

\bibitem[Khosla et~al.(2011)Khosla, Jayadevaprakash, Yao, and
  Fei-Fei]{KhoslaYaoJayadevaprakashFeiFei_FGVC2011}
Aditya Khosla, Nityananda Jayadevaprakash, Bangpeng Yao, and Li Fei-Fei.
\newblock Novel dataset for fine-grained image categorization.
\newblock In \emph{First Workshop on Fine-Grained Visual Categorization, IEEE
  Conference on Computer Vision and Pattern Recognition}, Colorado Springs, CO,
  2011.

\bibitem[Kim et~al.(2021)Kim, Kim, Seo, and Yoon]{kim2021xprotonet}
Eunji Kim, Siwon Kim, Minji Seo, and Sungroh Yoon.
\newblock Xprotonet: Diagnosis in chest radiography with global and local
  explanations.
\newblock In \emph{Proceedings of the IEEE/CVF Conference on Computer Vision
  and Pattern Recognition}, pages 15719--15728, 2021.

\bibitem[Krause et~al.(2013)Krause, Stark, Deng, and Fei-Fei]{6755945}
Jonathan Krause, Michael Stark, Jia Deng, and Li Fei-Fei.
\newblock 3d object representations for fine-grained categorization.
\newblock In \emph{2013 IEEE International Conference on Computer Vision
  Workshops}, pages 554--561, 2013.

\bibitem[Lapuschkin et~al.(2019)Lapuschkin, Wäldchen, Binder, Montavon, Samek,
  and Müller]{Lapuschkin_2019}
Sebastian Lapuschkin, Stephan Wäldchen, Alexander Binder, Grégoire Montavon,
  Wojciech Samek, and Klaus-Robert Müller.
\newblock Unmasking clever hans predictors and assessing what machines really
  learn.
\newblock \emph{Nature Communications}, 10\penalty0 (1), 2019.

\bibitem[Lin et~al.(2018)Lin, Goyal, Girshick, He, and
  Dollár]{lin2018focallossdenseobject}
Tsung-Yi Lin, Priya Goyal, Ross Girshick, Kaiming He, and Piotr Dollár.
\newblock Focal loss for dense object detection, 2018.

\bibitem[Liu et~al.(2022)Liu, Hu, Lin, Yao, Xie, Wei, Ning, Cao, Zhang, Dong,
  Wei, and Guo]{liu2022swintransformerv2scaling}
Ze Liu, Han Hu, Yutong Lin, Zhuliang Yao, Zhenda Xie, Yixuan Wei, Jia Ning, Yue
  Cao, Zheng Zhang, Li Dong, Furu Wei, and Baining Guo.
\newblock Swin transformer v2: Scaling up capacity and resolution, 2022.

\bibitem[Ma et~al.(2024)Ma, Donnelly, Liu, Vosoughi, Rudin, and
  Chen]{ma2024interpretable}
Chiyu Ma, Jon Donnelly, Wenjun Liu, Soroush Vosoughi, Cynthia Rudin, and
  Chaofan Chen.
\newblock Interpretable image classification with adaptive prototype-based
  vision transformers.
\newblock \emph{Advances in Neural Information Processing Systems},
  37:\penalty0 41447--41493, 2024.

\bibitem[Michalski et~al.(2025)Michalski, Wr{\'o}bel, Bontempelli, Lu{\'s}tyk,
  Kniejski, Teso, Passerini, Zieli{\'n}ski, and
  Rymarczyk]{michalski2025personalized}
Tomasz Michalski, Adam Wr{\'o}bel, Andrea Bontempelli, Jakub Lu{\'s}tyk,
  Mikolaj Kniejski, Stefano Teso, Andrea Passerini, Bartosz Zieli{\'n}ski, and
  Dawid Rymarczyk.
\newblock Personalized interpretability--interactive alignment of prototypical
  parts networks.
\newblock \emph{arXiv preprint arXiv:2506.05533}, 2025.

\bibitem[Molnar(2025)]{molnar2025}
Christoph Molnar.
\newblock \emph{Interpretable Machine Learning}.
\newblock 3 edition, 2025.

\bibitem[Nauta et~al.(2021{\natexlab{a}})Nauta, Jutte, Provoost, and
  Seifert]{nauta2021looks}
Meike Nauta, Annemarie Jutte, Jesper Provoost, and Christin Seifert.
\newblock This looks like that, because... explaining prototypes for
  interpretable image recognition.
\newblock In \emph{Joint European Conference on Machine Learning and Knowledge
  Discovery in Databases}, pages 441--456. Springer, 2021{\natexlab{a}}.

\bibitem[Nauta et~al.(2021{\natexlab{b}})Nauta, Van~Bree, and
  Seifert]{nauta2021neural}
Meike Nauta, Ron Van~Bree, and Christin Seifert.
\newblock Neural prototype trees for interpretable fine-grained image
  recognition.
\newblock In \emph{Proceedings of the IEEE/CVF conference on computer vision
  and pattern recognition}, pages 14933--14943, 2021{\natexlab{b}}.

\bibitem[Pach et~al.(2025)Pach, Rymarczyk, Lewandowska, Tabor, and
  Zieli{\'n}ski]{pach2024lucidppn}
Mateusz Pach, Dawid Rymarczyk, Koryna Lewandowska, Jacek Tabor, and Bartosz
  Zieli{\'n}ski.
\newblock Lucidppn: Unambiguous prototypical parts network for user-centric
  interpretable computer vision.
\newblock \emph{ICLR}, 2025.

\bibitem[Rudin(2019)]{rudin2019stop}
Cynthia Rudin.
\newblock Stop explaining black box machine learning models for high stakes
  decisions and use interpretable models instead.
\newblock \emph{Nature machine intelligence}, 1\penalty0 (5):\penalty0
  206--215, 2019.

\bibitem[Rudin et~al.(2021)Rudin, Chen, Chen, Huang, Semenova, and
  Zhong]{rudin2021interpretablemachinelearningfundamental}
Cynthia Rudin, Chaofan Chen, Zhi Chen, Haiyang Huang, Lesia Semenova, and Chudi
  Zhong.
\newblock Interpretable machine learning: Fundamental principles and 10 grand
  challenges, 2021.

\bibitem[Russakovsky et~al.(2015)Russakovsky, Deng, Su, Krause, Satheesh, Ma,
  Huang, Karpathy, Khosla, Bernstein, Berg, and
  Fei-Fei]{russakovsky2015imagenetlargescalevisual}
Olga Russakovsky, Jia Deng, Hao Su, Jonathan Krause, Sanjeev Satheesh, Sean Ma,
  Zhiheng Huang, Andrej Karpathy, Aditya Khosla, Michael Bernstein,
  Alexander~C. Berg, and Li Fei-Fei.
\newblock Imagenet large scale visual recognition challenge, 2015.

\bibitem[Rymarczyk et~al.(2021)Rymarczyk, Struski, Tabor, and
  Zieli{\'n}ski]{rymarczyk2021protopshare}
Dawid Rymarczyk, {\L}ukasz Struski, Jacek Tabor, and Bartosz Zieli{\'n}ski.
\newblock Protopshare: Prototypical parts sharing for similarity discovery in
  interpretable image classification.
\newblock In \emph{Proceedings of the 27th ACM SIGKDD Conference on Knowledge
  Discovery \& Data Mining}, pages 1420--1430, 2021.

\bibitem[Rymarczyk et~al.(2022{\natexlab{a}})Rymarczyk, Kaczy{\'n}ska, Kraus,
  Pardyl, and Zieli{\'n}ski]{rymarczyk2022protomil}
Dawid Rymarczyk, Aneta Kaczy{\'n}ska, Jaros{\l}aw Kraus, Adam Pardyl, and
  Bartosz Zieli{\'n}ski.
\newblock Protomil: Multiple instance learning with prototypical parts for
  fine-grained interpretability.
\newblock In \emph{Joint European Conference on Machine Learning and Knowledge
  Discovery in Databases}. Springer, 2022{\natexlab{a}}.

\bibitem[Rymarczyk et~al.(2022{\natexlab{b}})Rymarczyk, Struski, G{\'o}rszczak,
  Lewandowska, Tabor, and Zieli{\'n}ski]{rymarczyk2022interpretable}
Dawid Rymarczyk, {\L}ukasz Struski, Micha{\l} G{\'o}rszczak, Koryna
  Lewandowska, Jacek Tabor, and Bartosz Zieli{\'n}ski.
\newblock Interpretable image classification with differentiable prototypes
  assignment.
\newblock In \emph{European Conference on Computer Vision}, pages 351--368.
  Springer, 2022{\natexlab{b}}.

\bibitem[Rymarczyk et~al.(2023{\natexlab{a}})Rymarczyk, Dobrowolski, and
  Danel]{rymarczyk2022progrest}
Dawid Rymarczyk, Daniel Dobrowolski, and Tomasz Danel.
\newblock Progrest: Prototypical graph regression soft trees for molecular
  property prediction.
\newblock \emph{SIAM International Conference on Data Mining},
  2023{\natexlab{a}}.

\bibitem[Rymarczyk et~al.(2023{\natexlab{b}})Rymarczyk, van~de Weijer,
  Zieli{\'n}ski, and Twardowski]{rymarczyk2023icicle}
Dawid Rymarczyk, Joost van~de Weijer, Bartosz Zieli{\'n}ski, and Bartlomiej
  Twardowski.
\newblock Icicle: Interpretable class incremental continual learning.
\newblock In \emph{Proceedings of the IEEE/CVF International Conference on
  Computer Vision}, pages 1887--1898, 2023{\natexlab{b}}.

\bibitem[Sacha et~al.(2023)Sacha, Rymarczyk, Struski, Tabor, and
  Zieliński]{sacha2023protoseg}
Mikołaj Sacha, Dawid Rymarczyk, Lukasz Struski, Jacek Tabor, and Bartosz
  Zieliński.
\newblock Protoseg: Interpretable semantic segmentation with prototypical
  parts.
\newblock In \emph{Winter Conference on Applications of Computer Vision
  (WACV)}, 2023.

\bibitem[Sacha et~al.(2024)Sacha, Jura, Rymarczyk, Struski, Tabor, and
  Zieli{\'n}ski]{sacha}
Miko{\l}aj Sacha, Bartosz Jura, Dawid Rymarczyk, {\L}ukasz Struski, Jacek
  Tabor, and Bartosz Zieli{\'n}ski.
\newblock Interpretability benchmark for evaluating spatial misalignment of
  prototypical parts explanations.
\newblock In \emph{Proceedings of the AAAI Conference on Artificial
  Intelligence}, pages 21563--21573, 2024.

\bibitem[Struski et~al.(2024)Struski, Rymarczyk, and
  Tabor]{struski2024infodisent}
{\L}ukasz Struski, Dawid Rymarczyk, and Jacek Tabor.
\newblock Infodisent: Explainability of image classification models by
  information disentanglement.
\newblock \emph{arXiv preprint arXiv:2409.10329}, 2024.

\bibitem[Sundararajan et~al.(2017)Sundararajan, Taly, and
  Yan]{sundararajan2017axiomaticattributiondeepnetworks}
Mukund Sundararajan, Ankur Taly, and Qiqi Yan.
\newblock Axiomatic attribution for deep networks, 2017.

\bibitem[Ukai et~al.(2022)Ukai, Hirakawa, Yamashita, and
  Fujiyoshi]{ukai2022looks}
Yuki Ukai, Tsubasa Hirakawa, Takayoshi Yamashita, and Hironobu Fujiyoshi.
\newblock This looks like it rather than that: Protoknn for similarity-based
  classifiers.
\newblock In \emph{The Eleventh International Conference on Learning
  Representations}, 2022.

\bibitem[Wah et~al.(2023)Wah, Branson, Welinder, Perona, and
  Belongie]{wah_2023_cvm3y-5hh21}
Catherine Wah, Steve Branson, Peter Welinder, Pietro Perona, and Serge
  Belongie.
\newblock The caltech-ucsd birds-200-2011 dataset, 2023.

\bibitem[Wang et~al.(2021)]{wang2021interpretable}
Jiaqi Wang et~al.
\newblock Interpretable image recognition by constructing transparent embedding
  space.
\newblock In \emph{Proceedings of the IEEE/CVF International Conference on
  Computer Vision}, pages 895--904, 2021.

\bibitem[Xue et~al.(2022)Xue, Huang, Zhang, Cheng, Song, Wu, and
  Song]{xue2022protopformerconcentratingprototypicalparts}
Mengqi Xue, Qihan Huang, Haofei Zhang, Lechao Cheng, Jie Song, Minghui Wu, and
  Mingli Song.
\newblock Protopformer: Concentrating on prototypical parts in vision
  transformers for interpretable image recognition, 2022.

\bibitem[Zhang et~al.(2022)Zhang, Liu, Wang, Lu, and Lee]{zhang2021protgnn}
Zaixi Zhang, Qi Liu, Hao Wang, Chengqiang Lu, and Cheekong Lee.
\newblock Protgnn: Towards self-explaining graph neural networks.
\newblock 2022.

\end{thebibliography}
}



\end{document}